\documentclass{article} 
\usepackage[preprint, nonatbib]{neurips_2022}  
\usepackage{color}
\usepackage{amsmath,amsfonts,amssymb,amsthm}
\usepackage{cleveref}
\usepackage{subfig}
\usepackage{pgfplots}
\usepackage{pgfplotstable}
\pgfplotsset{compat=newest}
\usepackage{filecontents}
\usepackage{multirow}
\usepackage[flushleft]{threeparttable}
\usepackage{bm}
\usepackage{booktabs}             
\usepackage{textcomp}             
\usepackage{balance}

\renewcommand{\vec}[1]{\boldsymbol{#1}}
\renewcommand{\paragraph}[1]{\textit{#1}}

\definecolor{myorange}{RGB}{238,97,42}  %
\definecolor{myblue}{RGB}{178,179,249}  
\definecolor{mygrey}{RGB}{166,166,166}  %
\definecolor{mygreen}{RGB}{180,210,36}  
\definecolor{myred}{RGB}{238,0,0}       
\definecolor{myyellow}{RGB}{198,148,34} 
\definecolor{mydark}{RGB}{114,44,114}   
\definecolor{mymiddle}{RGB}{144,44,144} 
\definecolor{mylight}{RGB}{167,44,167}  

\usepackage{times}  
\usepackage{helvet}  
\usepackage{courier}  
\usepackage[hyphens]{url}  
\usepackage{graphicx} 
\usepackage[numbers]{natbib}  
\usepackage{caption} 
\usepackage{bbm}
\usepackage{comment}
\usepackage{algorithm}
\usepackage{algpseudocode}
\graphicspath{{figs/}}
\usepackage{newfloat}
\usepackage{listings}
\DeclareCaptionStyle{ruled}{labelfont=normalfont,labelsep=colon,strut=off} 
\floatstyle{ruled}
\newfloat{listing}{tb}{lst}{}
\floatname{listing}{Listing}
\title{An Adversarial Active Sampling-based Data Augmentation Framework for Manufacturable Chip Design}

\author{
Mingjie Liu$^{1,2}$ \quad Haoyu Yang$^{1}$ \quad Zongyi Li$^{1,3}$ \quad Kumara Sastry$^1$ \quad Saumyadip Mukhopadhyay$^1$ \\ 
\textbf{Selim Dogru}$^1$ \quad \textbf{Anima Anandkumar}$^{1,3}$ \quad \textbf{David Z. Pan}$^{2}$ \quad \textbf{Brucek Khailany}$^{1}$ \quad \textbf{Haoxing Ren}$^{1}$ \\
$^1$ Nvidia Corporation \quad $^2$ The University of Texas at Austin \quad $^3$ California Institute of Technology \\
\texttt{jay\_liu@utexas.edu}, \texttt{haoxingr@nvidia.com}
}

\begin{document}

\maketitle

\begin{abstract}
Lithography modeling is a crucial problem in chip design to ensure a chip design mask is manufacturable. 
It requires rigorous simulations of optical and chemical models that are computationally expensive. 
Recent developments in machine learning have provided alternative solutions in replacing the time-consuming lithography simulations with deep neural networks. 
However, the considerable accuracy drop still impedes its industrial adoption. 
Most importantly, the quality and quantity of the training dataset directly affect the model performance.
To tackle this problem, we propose a litho-aware data augmentation (LADA) framework to resolve the dilemma of limited data and improve the machine learning model performance. 
First, we pretrain the neural networks for lithography modeling and a gradient-friendly StyleGAN2 generator.
We then perform \textit{adversarial active sampling} to generate informative and synthetic in-distribution mask designs.
These synthetic mask images will augment the original limited training dataset used to finetune the lithography model for improved performance.
Experimental results demonstrate that LADA can successfully exploits the neural network capacity by narrowing down the performance gap between the training and testing data instances.
\end{abstract}

\section{Introduction}
The advancement of semiconductor industry has enabled rapid development of AI and deep learning technologies,
which in turn offers great opportunities for novel chip design methodology, enabling faster design turn-around-time, better PPA (power, performance and area) and higher yield \cite{googleplace,PD-DAC2021-Ren,OPC-TCAD2020-Yang,PLACE-NIPS2021-Cheng,PLACE-DAC2019-DREAMPlace}.
Particularly, recent researches have demonstrated efficacy using reinforcement learning to place circuit components (macros and standard cells) on to chip canvas \cite{googleplace,PLACE-NIPS2021-Cheng}, which is one of the time consuming phases in chip design flow.

In this paper, we focus on the lithography modeling, a critical problem in chip design and manufacturing flow, which has been a very active area for machine learning applications since 2010s \cite{HSD-ASPDAC2011-Ding,HSD-SPIE2012-Matsunawa,DFM-SPIE2017-Shim}. Lithography modeling computes the patterns (resist image) of a chip design (mask image) on the silicon wafer without going through real manufacturing process.
It allows designers to find potential circuit manufacture failures and conduct necessary post design optimization prior to manufacturing.
Lithography modeling traditionally consists of two stages \cite{DFM-B2011-Ma}. 
The optical modeling outputs the intensity (aerial image) of the light beams that are projected on the silicon wafer, 
which is given by the weighted convolution between the mask image and a set of lithography kernels. 
The resist modeling then applies thresholds on the intensity map and obtains the final resist image on the wafer.

However, legacy approaches are time consuming to compute complex physical models, normally taking couple of seconds to minutes. 
To reduce the design turn-around-time, recent researchers have investigated the possibility of using deep neural network models as an alternative of physical models, reducing the computing runtime by nearly 3 orders of magnitudes. 
Commonly used model architectures are UNet-backboned image-to-image translation networks \cite{UNet,OPC-ICCAD2020-DAMO,DFM-ISPD2020-Ye,DFM-DAC2019-Ye},
which can make fast resist image predictions with moderate accuracy loss.  State-of-art dual-band optics-inspired neural networks (DOINN) 
\cite{DFM-DAC2022-Yang} further integrates lithography physics into the neural network design and improves the prediction accuracy. 
However the quality gap between the machine predicted resist image and the real physical simulated results still prohibits learning-based solutions in the chip manufacturing flow.
Resolving this challenge is not trivial because of the limited availability of high quality data.

\begin{figure*}[tb!]
	\centering
	\includegraphics[width=.72\textwidth]{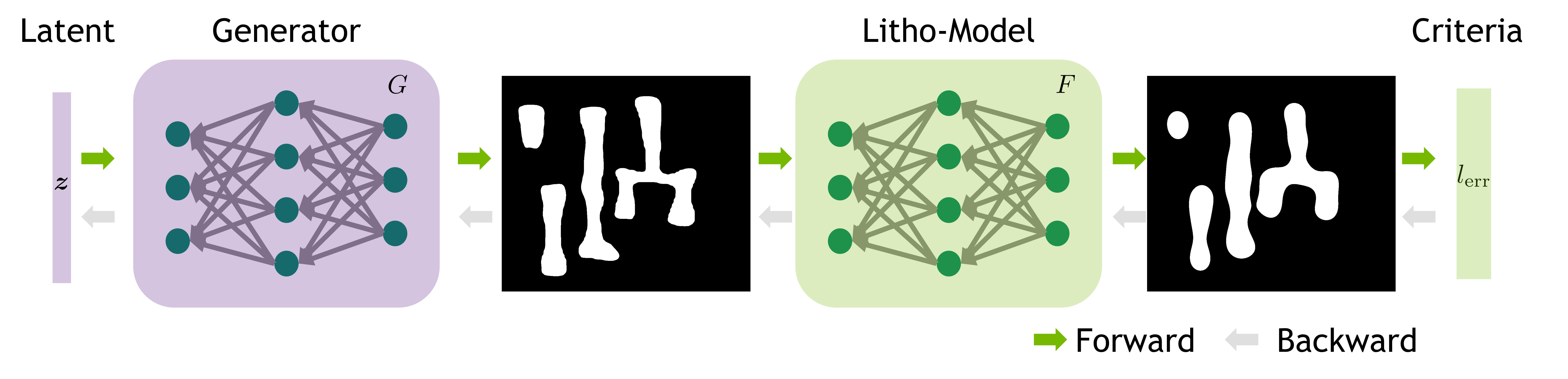}
	\caption{Overall framework of LADA.}
	\label{fig:lada}
\end{figure*}

To address data limitation, we present a litho-aware data-augmentation framework (LADA) that takes advantage of adversarial data generation~\cite{GAN-data-augmentation, GAN-data-augmentation2}, active learning~\cite{active-learning-survey,GAAL}, and modern generative adversarial network (GAN) design (\Cref{fig:lada}).  
Specifically we want to generate realistic in-distribution data with increased data efficiency, such that the synthesized data are most informative for model improvement. 
LADA is designed with an end-to-end data framework where a pretrained GAN generator and a lithography modeling network are connected.
The generator enforces in-distribution data generation, and the generator latent are further optimized based on sampling criteria from the model output to generate informative examples most likely to fail predictions. 

\section{Method}
We present our litho-aware data augmentation (LADA) framework, where we use StyleGAN2 to generate realistic in-distribution data and formulate \textit{adversarial active sampling} as an optimization problem to harness informative samples based on model output.

\Cref{fig:lada} illustrates the pipeline architecture of LADA, where the generator $G$ targets to generate mask images that will fail the lithography modeling network $F$. 
For the best of LADA performance, we select StyleGAN-2~\cite{stylegan2} as our generator backbone and DOINN \cite{DFM-DAC2022-Yang} to be the machine learning-based lithography model.
However, their limitations require litho-dedicated design to make the whole framework feasible.

We leverage StyleGAN2 generator to synthesize realistic in-distribution data.
StyleGAN2 is a state-of-the-art generative model trained in adversarial setting, where a novel generator architecture leads to disentangled high-level attributes and stochastic variations in an unsupervised manner. 
The generated \emph{styles} variables from random \emph{latent} code control granular style variations of the generated image, such as the background lightning or sex of human faces.
The random \emph{noise} variables affect low-level details of the images, allowing small perturbations barely noticeable to the human eye, injecting stochastic details such as finer details background details or finer curls of hair.
This attribute is consistent with our demands on mask image generation because the lithography process is highly sensitive and minor perturbations on the mask image will result in significant change on the output.

\begin{figure}[tb!]
	\centering
	\includegraphics[width=.45\textwidth]{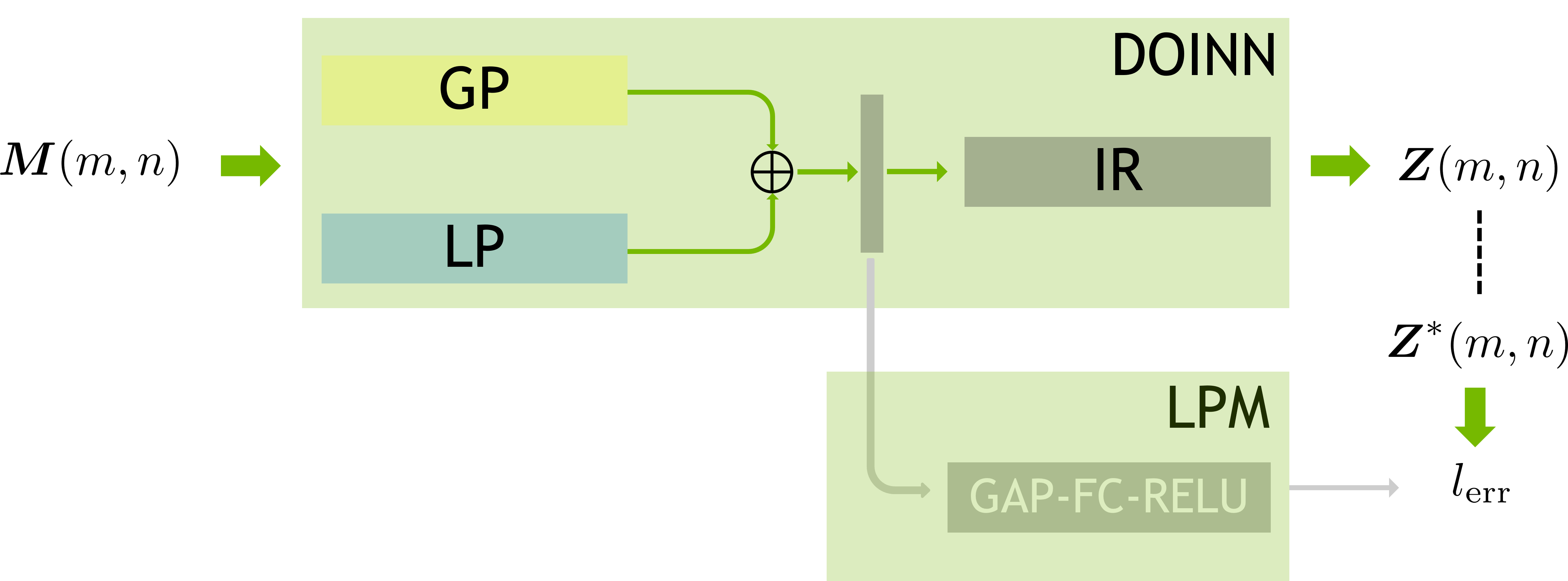}
	\caption{Multi-task DOINN with loss prediction module.}
	\label{fig:doinn}
\end{figure}

As shown in \Cref{fig:doinn}, the DOINN backbone consists of three processing paths: global perception (GP), local perception (LP) and image reconstruction (IR). GP and LP work together to capture high quality embedding that will be feed into IR to generate resist images.
Inspired by Fourier Optics, the GP path leverage Fourier Neural Operator~\cite{PDE-ICLR2021-Li} to obtain low-frequency global information feature maps, while the LP path consists of a series of convolutional layers for high frequency local information.
The obtained bottleneck feature maps are fed to a series of deconvolution layers for image reconstruction, similar to a U-Net~\cite{UNet} structure.
To make the LADA framework feasible, we leverage the \textit{loss prediction module}~\cite{loss_prediction_module} that can be integrated into the original DOINN following GP and LP. 

\paragraph{Loss Prediction Module:}
The loss prediction module (LPM) predict the loss value $\hat{l_{x_U}}$ without ground truth labels for any given unlabeled $x_{U}$ in the input domain.
It takes feature maps as the inputs that are extracted between mid-level blocks of the target DOINN model.
Each feature map is reduced to a fixed embedding size through the LPM module, which consists sequentially of global average pooling (GAP), FC layers, and ReLU activation layers.
The embeddings from different layers are concatenated and finally passed through another FC layer to predict the scalar loss value.
    \label{eq:lpm_loss}

The goal of adversarial active sampling lies in two key aspects: (1) Generate realistic in-distribution synthetic data; and (2) Improve data efficiency and training overhead with generating most informative data inputs.
Prior work on active learning focus on reducing the labeling cost by selecting from pool-based settings, where our work synthesize images from a pre-trained GAN generator in membership query synthesis fashion.
The optimization problem that describe novel and informative sample generation is as follows:
\begin{equation}
    \underset{\vec{z}, \mathrm{\textit{noise}} \in N(0,1)}{\mathrm{argmax}} C( G(\vec{z},\textit{noise}; \Theta_{G}), F(\cdot; \Theta_{F,t}) ),
    \label{eq:active_learning}
\end{equation}
where $\vec{z}, \textit{noise}$ are the latent code and \textit{noise} variables, $G$ is the StyleGAN2 generator with pre-trained weights $\Theta_{G}$, and $F$ is the DOINN model with weights $\Theta_{F,t}$ during some iteration $t$.
$C$ is a criteria function to evaluate how informative the synthesize image would be towards at improving the model $F$ if the model is to be retrained with the image added at the next iteration $t+1$.

We explore two different solutions to the relaxed original problem, where the sampling process is conducted on separated input domains of the StyleGAN2 generator \textit{style} latent code $\vec{z}$ and \textit{noise}.
Sampling in the \textit{style} latent code $\vec{z}$ domain is equivalent to:
\begin{equation}
\begin{split}
  \underset{\vec{z} }{\mathrm{argmax}}C_{\vec{z}} ( G(\vec{z}; \Theta_{G}), F(\cdot; \Theta_{F,t})  ) \\
  + \frac{\lambda_{1}}{|\vec{z}|} \mathrm{log}(p(\vec{z}; N(0,1)) ,
 \end{split}
 \label{eq:adv_style}
\end{equation}
where the constraint that $\vec{z} \in N(0,1)$ is relaxed instead to maximize the log-likelihood and \textit{noise} is fixed to 0.
Similarly, we can random sample $\vec{z}$ from $N(0,1)$ which is kept fixed and optimize \textit{noise} variable: 
\begin{equation}
\begin{split}
  \underset{\mathrm{\textit{noise}} }{\mathrm{argmax}}C_{\mathrm{\textit{noise}}} ( G(\mathrm{\textit{noise}}; \Theta_{G}), F(\cdot; \Theta_{F,t})  ) \\
  + \frac{\lambda_{2}}{|N|} \mathrm{log}(p(\mathrm{\textit{noise}}; N(0,1)).
 \end{split}
 \label{eq:adv_noise}
\end{equation}

We leverage the multi-task DOINN LPM predicted loss as the sampling criteria:
\begin{equation}
    C=F_{LPM}(G(\vec{z}, \mathrm{\textit{noise}})),
\end{equation}
which motivates synthesized data to have larger loss between DOINN model $F$ prediction and labels without having to access the lithography simulator.
This method is domain agnostic as the network learns a single loss scalar and should generalize to synthesized image regardless of granular \textit{style} changes or small \textit{noise} perturbations.
The output of the generator $G$ is directly chained as the input of the DOINN LPM model $F_{LPM}$.

We also experiment with alternative sampling criterion from methods in pool-based active learning and adversarial attack.

\paragraph{Model Uncertainty:}
The Dice Loss~\cite{DICE} between prediction logits $\{p_i, q_i\}$ of the DOINN model $F$ captures the model uncertainty:
\begin{equation}
    C = L_{Dice}(P,Q) - 1 = -\frac{2\sum p_i q_i}{\sum p_i^2 + \sum q_i^2}.
\end{equation}

\paragraph{Model Perturbation:}
Inspired by adversarial attack~\cite{adv_attack_segmentation}, the cross-entropy loss between model $F$ outputs samples maximal model perturbations:
\begin{equation}
    C_{\mathrm{\textit{noise}}} = L_{CE} (F(G(\vec{z}, \mathrm{\textit{noise}})), \tilde{F}(G(\vec{z}, 0))).
\end{equation}
\section{Experiments}

Lithography modeling lacks sufficient open-source datasets to train deep machine learning models.
In this section, we briefly explain how we constructed the initial training dataset, and pretrain models for StyleGAN2 and DOINN.


\paragraph{Testing Dataset:}
We test our models on ICCAD 2013 CAD Contest~\cite{OPC-ICCAD2013-Banerjee} which consists of 10 M1 (Metal Layer 1) designs on $32nm$ design node at 2k resolution.

\paragraph{Initial Training Dataset:}
We follow prior work~\cite{OPC-DAC2018-Yang} in generating a \textit{shape}-based generator to synthesize initial training dataset.
The \textit{shape}-based generator synthesized rectlinear design patterns following design rules similar to test data patterns.
The generated images are optimized for Optical Proximity Correction (OPC) with industrial level tools, such that the mask images are manufacturable.
The synthesized mask images after OPC are simulated with a lithograph simulator~\cite{OPC-ICCAD2021-Chen} to obtain the resist image label.\footnote{OPC takes hours to complete for a single design, while lithography simulation takes seconds.}
We construct an initial training dataset of 2000 M1 designs for model pretraining.

\paragraph{Model Pretraining}
The StyleGAN2 and DOINN models are pretrained on the initial training dataset of 2000 M1 designs.
In training StyleGAN2\footnote{https://github.com/NVlabs/stylegan2-ada-pytorch} we follow the proposed default parameters.
The model is trained with downscaled image resolution of 256 to decrease training time and generated images are \textit{bilinear} upsampled to 2k resolution.

\paragraph{Evaluation Metrics}
We evaluate models using the Jacard index of the foreground class (\textit{fIoU}):
\begin{equation}
    fIoU = \frac{P_1 \cap G_1}{P_1 \cup G_1},
\end{equation}
where $P_1, G_1$ denotes the prediction and ground truth for the foreground class.
We use \textit{fIoU} since the high proportion of background pixels typically inflates the overall \textit{mIoU} score.

\begin{table}[ht!]
\centering
\caption{Testing results for iterative adversarial active sampling. The \textit{shape} use \textit{shape}-based pattern generator, \textit{random} randomly samples for \textit{style} and \textit{noise} from StyleGAN2, \textit{style\_dice} use dice loss for uncertainty in where we only sample for \textit{style} for granular changes in the generation, \textit{noise\_CE} is adversarial sampling on \textit{noise} domain for maximal model output perturbation, \textit{style\_pred} and \textit{noise\_pred} are adversarial sampling for the \textit{loss prediction module}'s predicted loss on \textit{style} and \textit{noise} respectively.
\textit{pretrain} presents both \textit{train} and \textit{test} results of the pretrained model on 2k data. Other results are models finetuned on all generated 32k data after 16 iterations.}
\begin{tabular}{c|ccc}
\toprule
Item                          & $fIoU\%$         & $error\%$       & $Gap\%$\\ \midrule
\textit{pretrain (train)}     & 98.4583          & 1.5417          & -\\
\textit{pretrain (test)}      & 94.3589          & 5.6411          & 4.0994\\ \midrule
\textit{shape}                & 96.3467          & 3.6533          & 2.1116\\ 
\textit{random}               & 97.1370          & 2.8630          & 1.3213\\ 
\textit{style\_dice}          & 97.3223          & 2.6777          & 1.1360\\ 
\textit{noise\_CE}            & 97.3222          & 2.6778          & 1.1361\\ 
\textbf{\textit{style\_pred}} & \textbf{98.2216} & \textbf{1.7784} & \textbf{0.2367}\\ 
\textit{noise\_pred}          & 98.1474          & 1.8526          & 0.3109 \\ \bottomrule
\end{tabular}
\label{tab:results}
\end{table}

We conduct experimental studies to evaluate the proposed active learning approach on improving lithograph modeling.
We extensively compare against different adversarial sampling criterion functions inspired from prior work on pool-based active learning.
We follow the proposed method in where we set total iteration to $T=16$ and the labeling budget for each iteration $B=2k$.
\Cref{tab:results} compares the testing results for iterative adversarial active learning.
\textit{Gap} measures the generalization gap between model performance with \textit{train} results on~\textit{pretrain}.
The \textit{style\_pred} performs the best, with a $fIoU$ testing error of $1.78\%$, reducing the generalization error by $37.9\%$ compared with \textit{random} of $2.86\%$, and narrows the generalization gap to less than $0.24\%$.

\begin{table}[tb!]
	\centering
	\caption{Mask legalization makes standard adversarial attack meaningless. Legalization restores noised injected mask image to the original image, yielding meaningless adversarial example generation. Adv masks are illegal and do not have PhysicalSim results.}
	\renewcommand{\arraystretch}{1.0}
	\begin{tabular}{m{.15\linewidth}|m{.22\linewidth}|m{.22\linewidth}|m{.22\linewidth}}
		\toprule
		Design               & Mask Image & ~~~~DOINN& PhysicalSim \\ \midrule
		Original & \includegraphics[width=.1\textwidth]{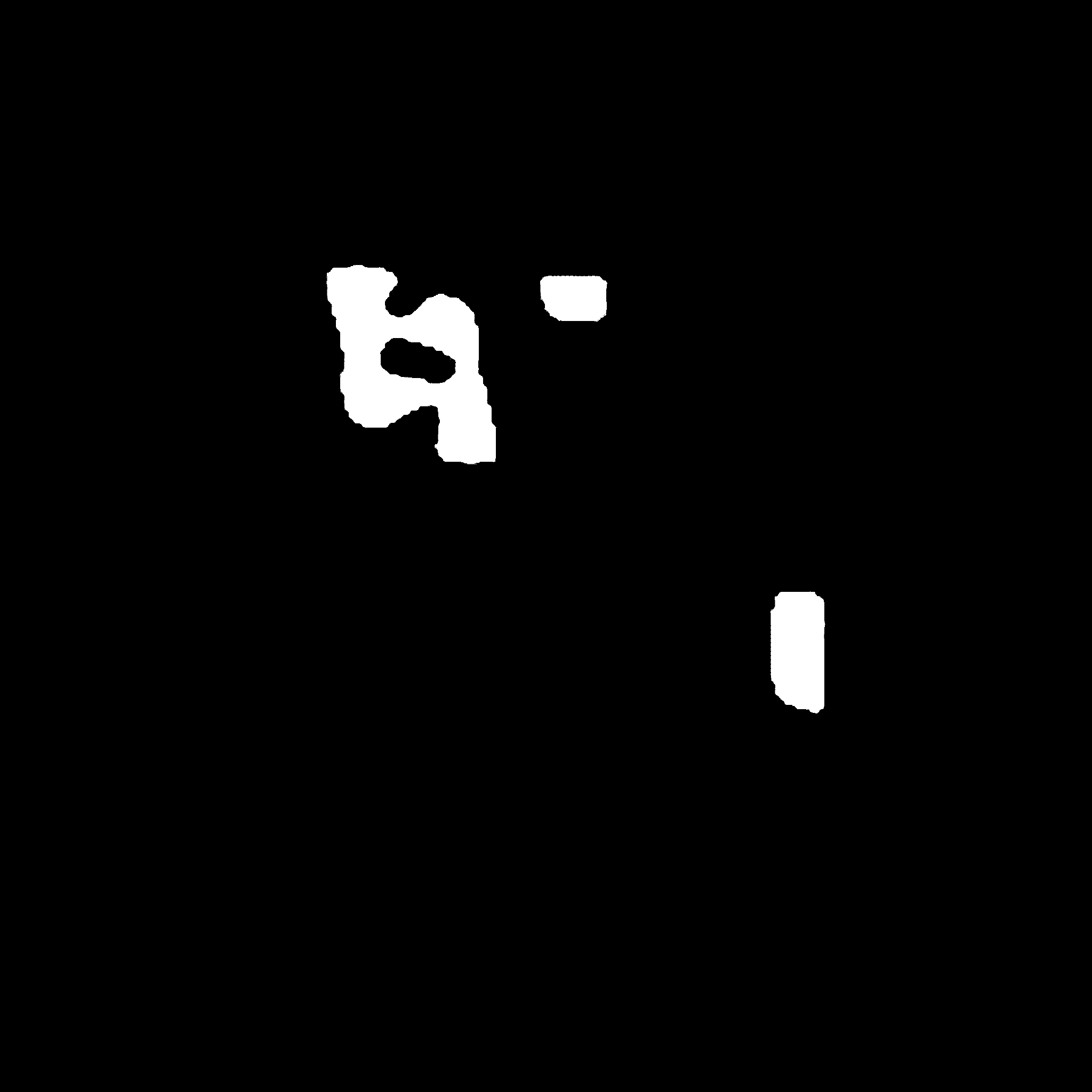}&\includegraphics[width=.1\textwidth]{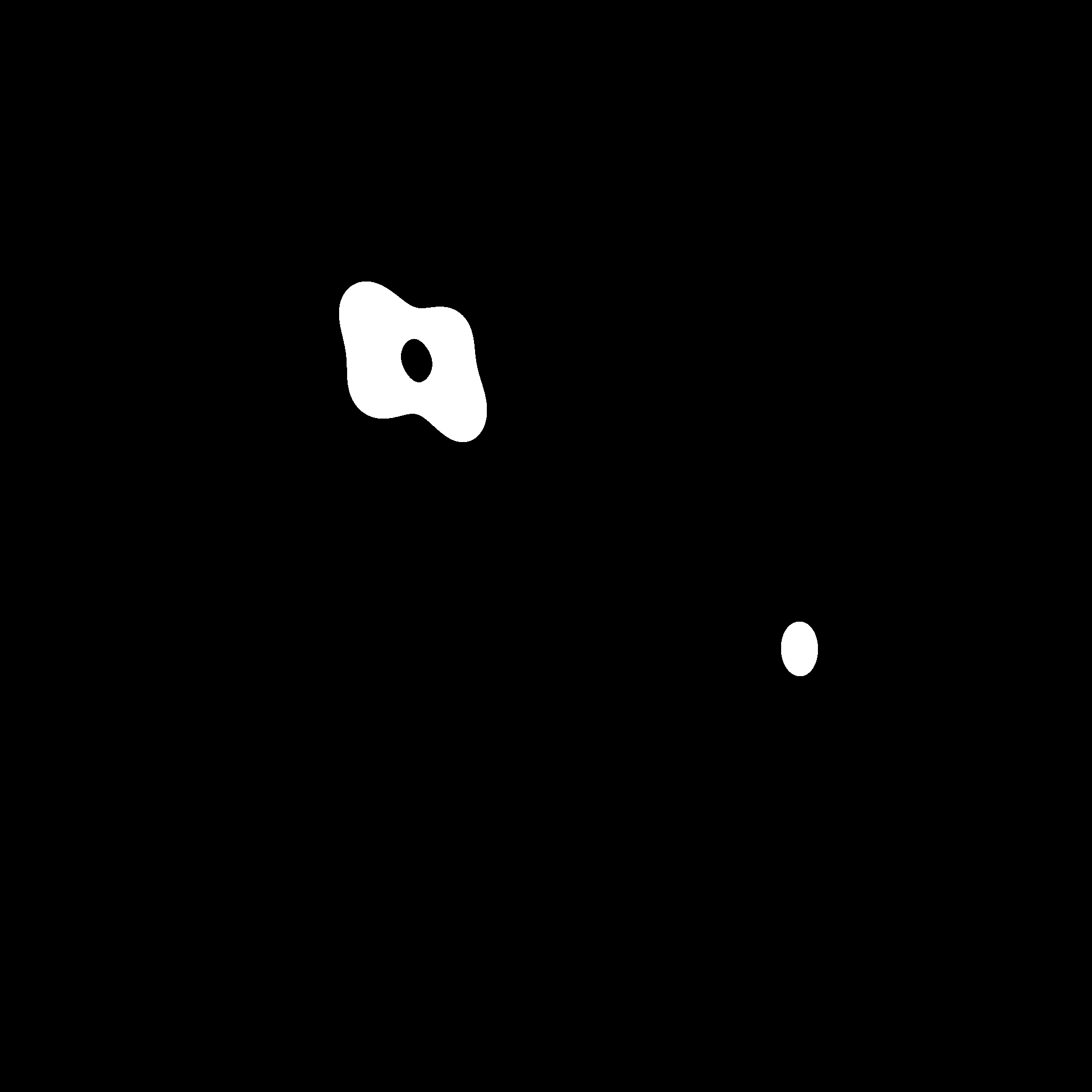}&\includegraphics[width=.1\textwidth]{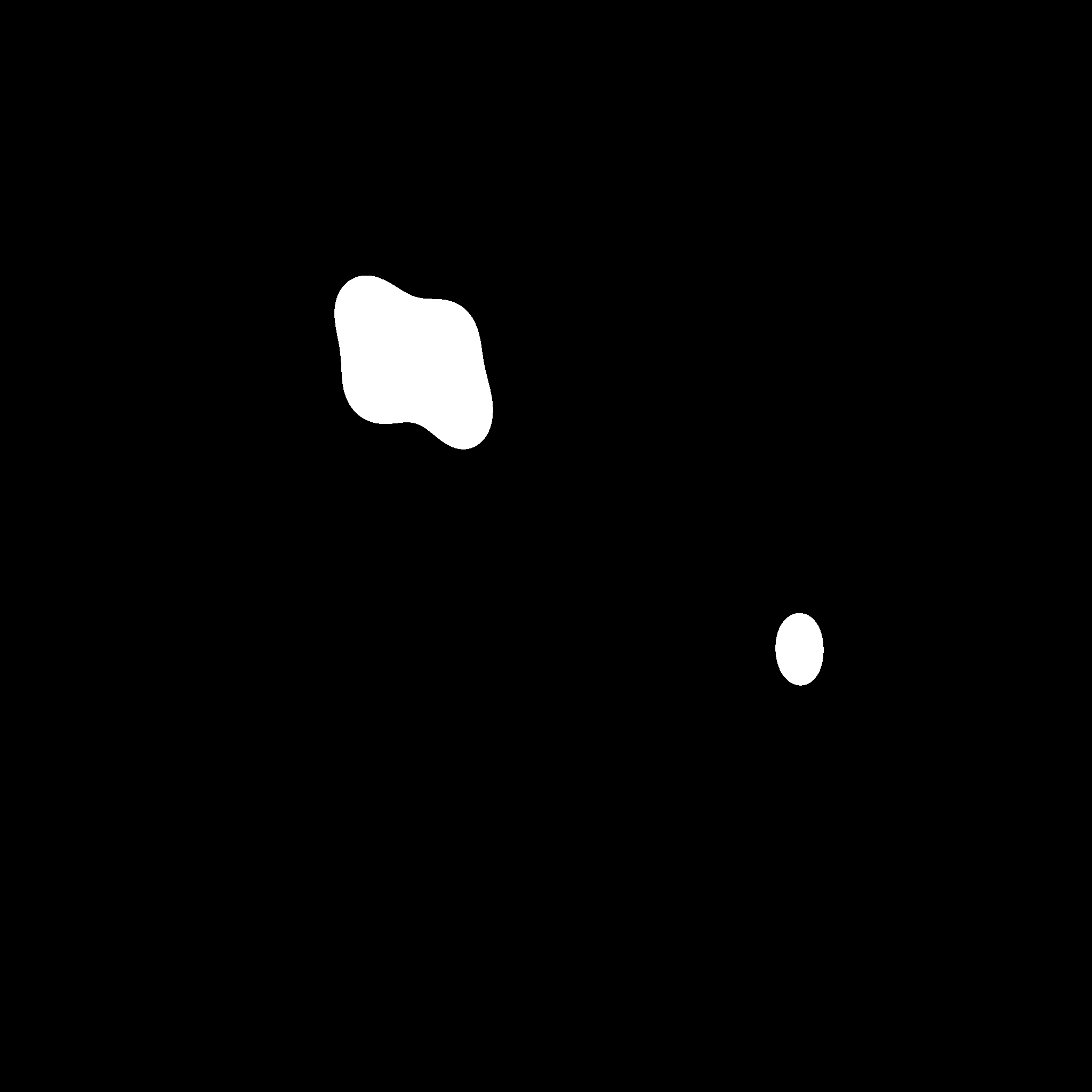}   \\ \midrule
		Adv      & \includegraphics[width=.1\textwidth]{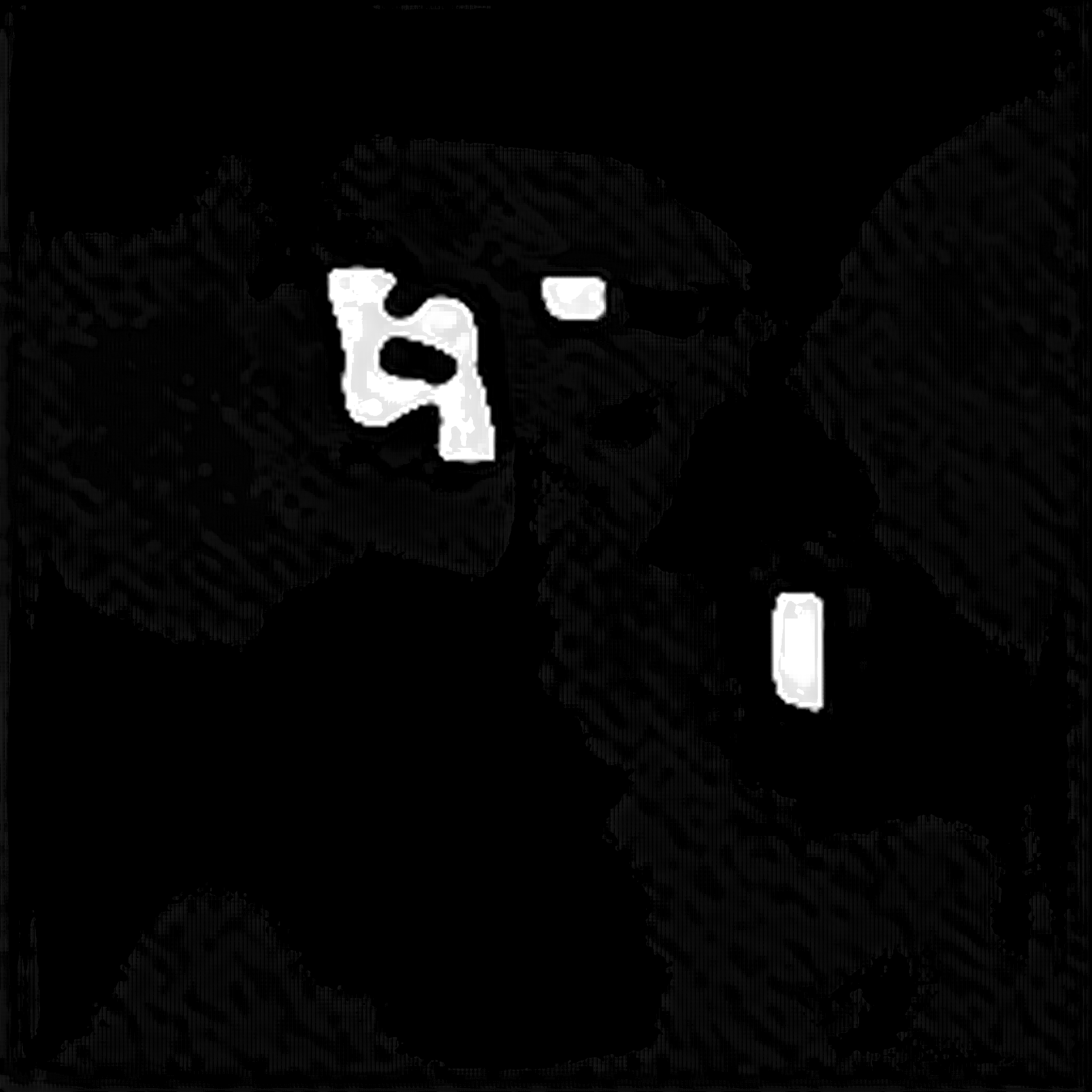}&\includegraphics[width=.1\textwidth]{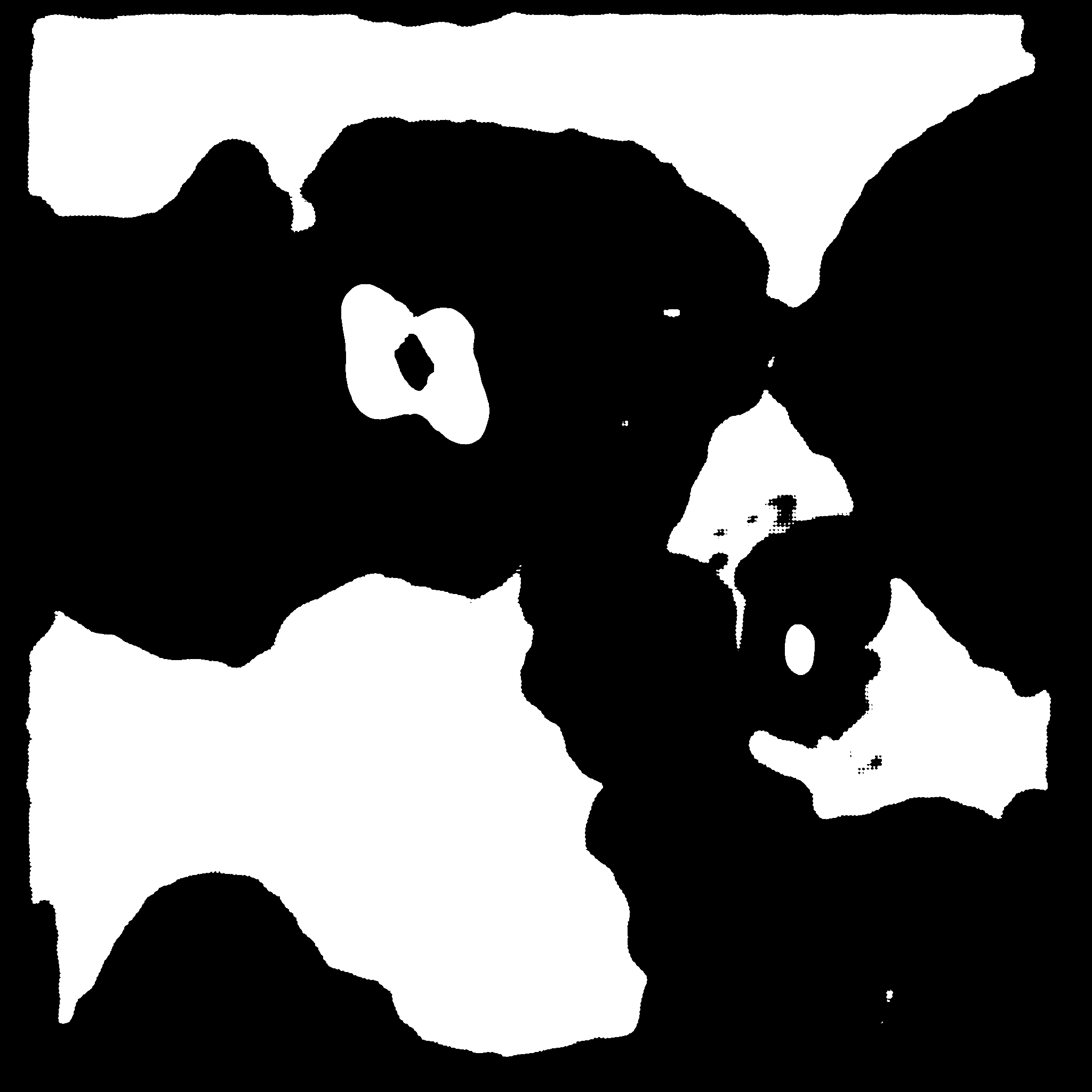}& {~~~~~~~N/A}  \\ \midrule
		Adv-Legal      & \includegraphics[width=.1\textwidth]{input_clean}  &\includegraphics[width=.1\textwidth]{model_output}&\includegraphics[width=.1\textwidth]{golden}    \\ \bottomrule
	\end{tabular}
	\label{tab:pixel_adv}
\end{table}

\begin{figure*}[tb!]
    \centering
    \includegraphics[width=0.85\textwidth]{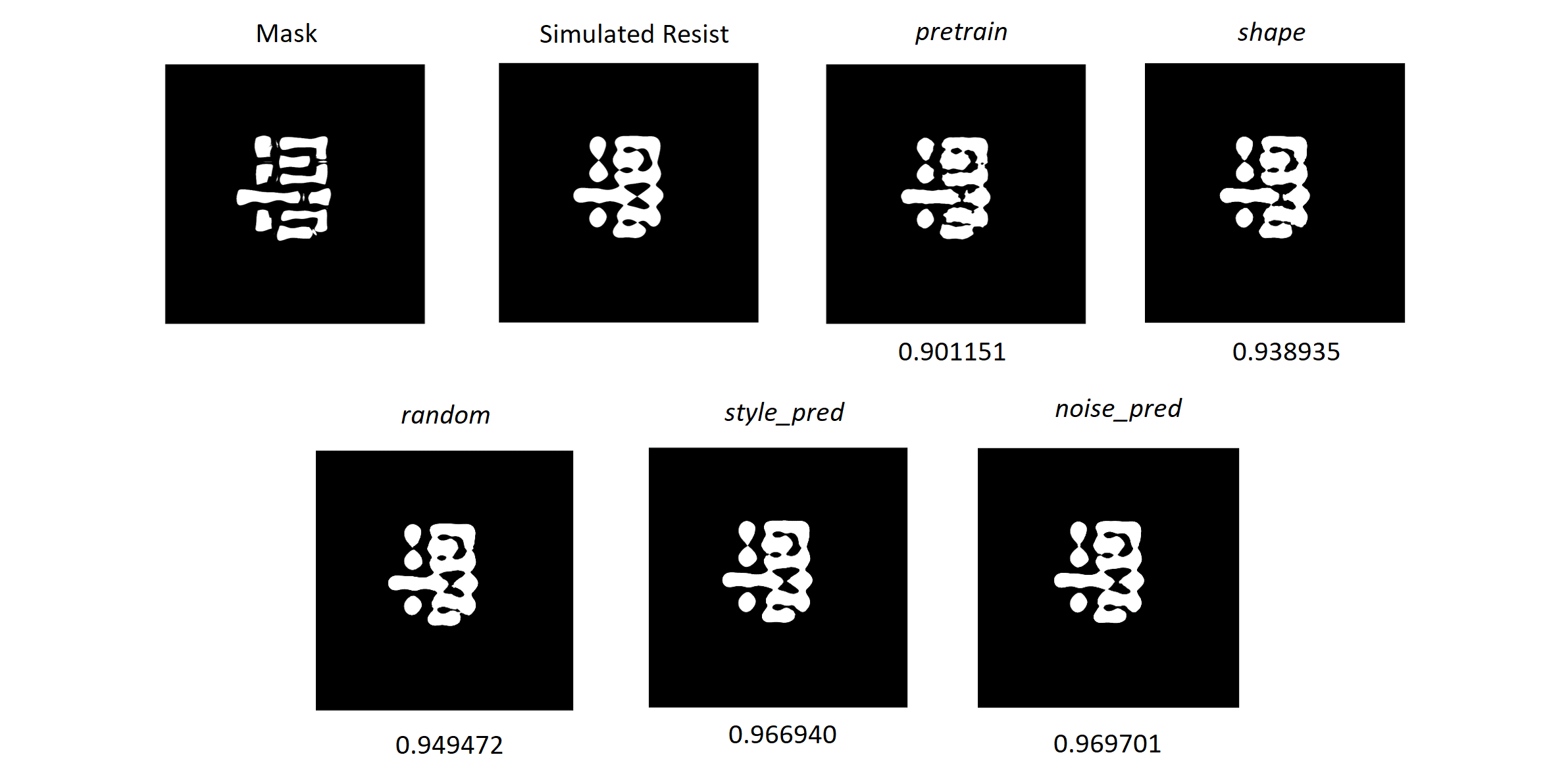} 
    \caption{Visualization and comparison for test performance. The $fIoU$ are listed at the bottom of each predictive output.}
\end{figure*}

\section{Conclusion and Future Work}
In this work, we present LADA, a litho-aware data augmentation framework based on adversarial active sampling techniques, which targets at reducing the gap between the empirical and the generalization error of the machine learning-based lithography simulator.
We aim at resolving both the data limitation in data synthesis while increasing the data efficiency and reducing training overhead through adversarial active sampling.
Our work in resolving data limitation and improving data quality in a query-based activation learning setting, could serve to further boost neural network generalization for learning physical-based systems.
Future work include further increasing the DOINN backbone model capacity and improving its generalization bound.

{
\bibliographystyle{abbrvnat}
\bibliography{ref/Top,ref/DFM,ref/HSD,ref/additional,ref/PD}
}

\end{document}